\documentclass{article}
\usepackage[english]{babel}
\usepackage[ruled, linesnumbered, noend]{algorithm2e}
\usepackage[small]{caption}
\usepackage[utf8]{inputenc}
\usepackage{amsfonts}
\usepackage{amssymb}
\usepackage{bm}
\usepackage{enumitem}
\usepackage{graphicx}
\usepackage{ijcai18}
\usepackage{mathrsfs}
\usepackage{multirow}
\usepackage{soul}
\usepackage{subfigure}
\usepackage{times}
\usepackage{xcolor}
\usepackage{tabu}
\usepackage{array}
\usepackage{booktabs}
\usepackage{siunitx}
\usepackage{amsmath} 
\usepackage[hidelinks]{hyperref}
\usepackage{subfigure}
\title{Constructing Narrative Event Evolutionary Graph for Script Event Prediction}

\author{
Zhongyang Li, Xiao Ding, Ting Liu\thanks{Corresponding author: tliu@ir.hit.edu.cn} \\ 
Research Center for Social Computing and Information Retrieval \\ Harbin Institute of Technology, China \\
\{zyli, xding, tliu\}@ir.hit.edu.cn
}

\begin{document}
\maketitle
\begin{abstract}
Script event prediction requires a model to predict the subsequent event given an existing event context. Previous models based on event pairs or event chains cannot make full use of dense event connections, which may limit their capability of event prediction. To remedy this, we propose constructing an event graph to better utilize the event network information for script event prediction. In particular, we first extract narrative event chains from large quantities of news corpus, and then construct a \textit{narrative event evolutionary graph} (NEEG) based on the extracted chains. NEEG can be seen as a knowledge base that describes event evolutionary principles and patterns. To solve the inference problem on NEEG, we present a \textit{scaled graph neural network} (SGNN) to model event interactions and learn better event representations. Instead of computing the representations on the whole graph, SGNN processes only the concerned nodes each time, which makes our model feasible to large-scale graphs. By comparing the similarity between input context event representations and candidate event representations, we can choose the most reasonable subsequent event. Experimental results on widely used New York Times corpus demonstrate that our model significantly outperforms state-of-the-art baseline methods, by using standard multiple choice narrative cloze evaluation.

\end{abstract}

\section{Introduction}
\begin{figure} \small
    \centering
    \includegraphics[width=0.8\columnwidth]{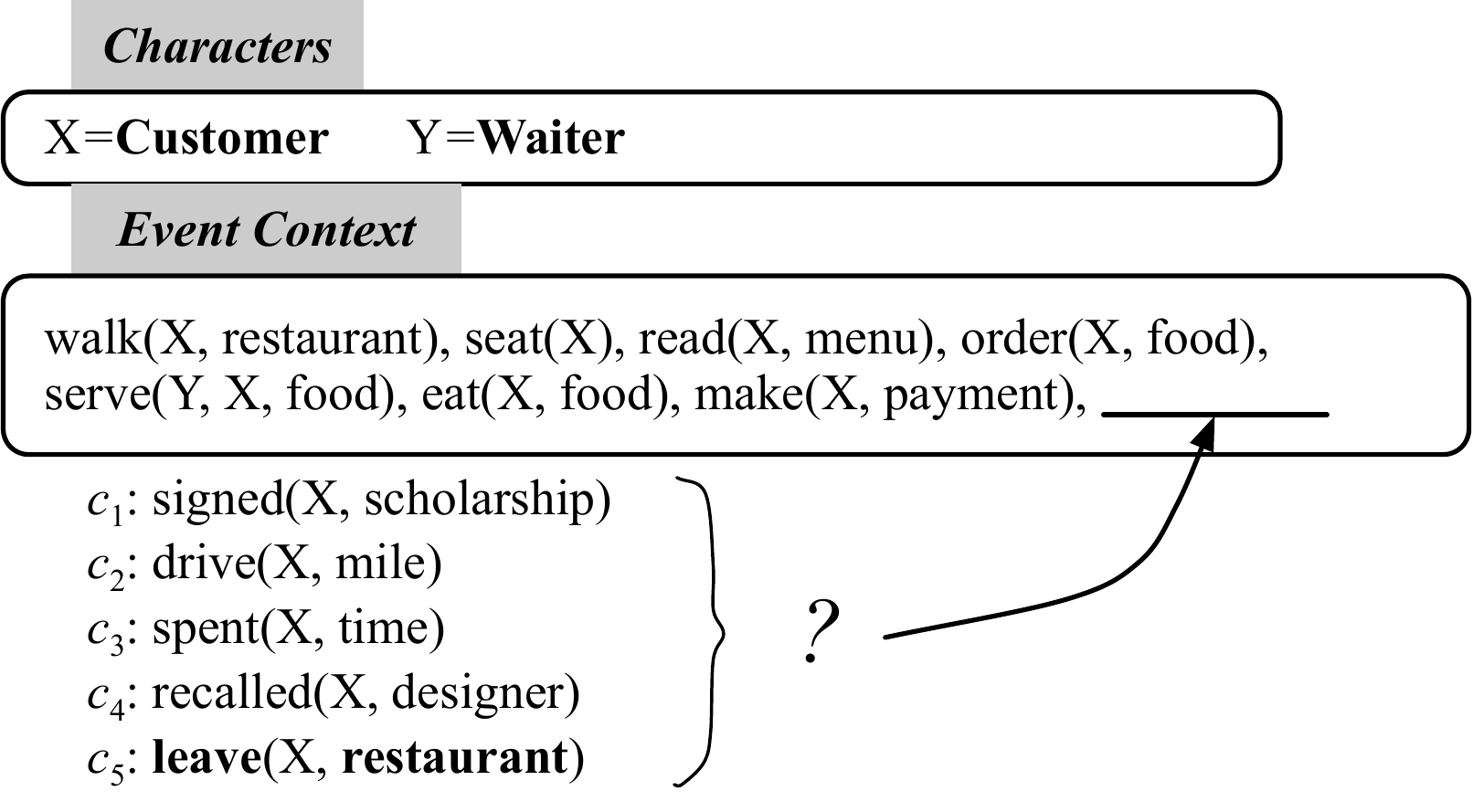}
    \vspace{-0.3cm}
    \caption{An example of script event prediction. The correct subsequent event is marked in bold, and other four choices are randomly selected.}
    \label{fig:task}
\end{figure}

Understanding events described in text is crucial to many artificial intelligence (AI) applications, such as discourse understanding, intention recognition and dialog generation. Script event prediction is the most challenging task in this line of work. This task was first proposed by \citeauthor{chambers2008} \shortcite{chambers2008}, who defined it as giving an existing event context, one needs to choose the most reasonable subsequent event from a candidate list (as shown in Figure \ref{fig:task}).

Previous studies built prediction models either based on event pairs \cite{chambers2008,MarkGW-AAAI16} or event chains \cite{wang2017integrating}. Although success in using event pairs and chains, rich connections among events are not fully explored. To better model event connections, we propose to solve the problem of script event prediction based on event graph structure and infer the correct subsequent event based on network embedding.

\begin{figure} 
    \centering
    \includegraphics[width=1\columnwidth]{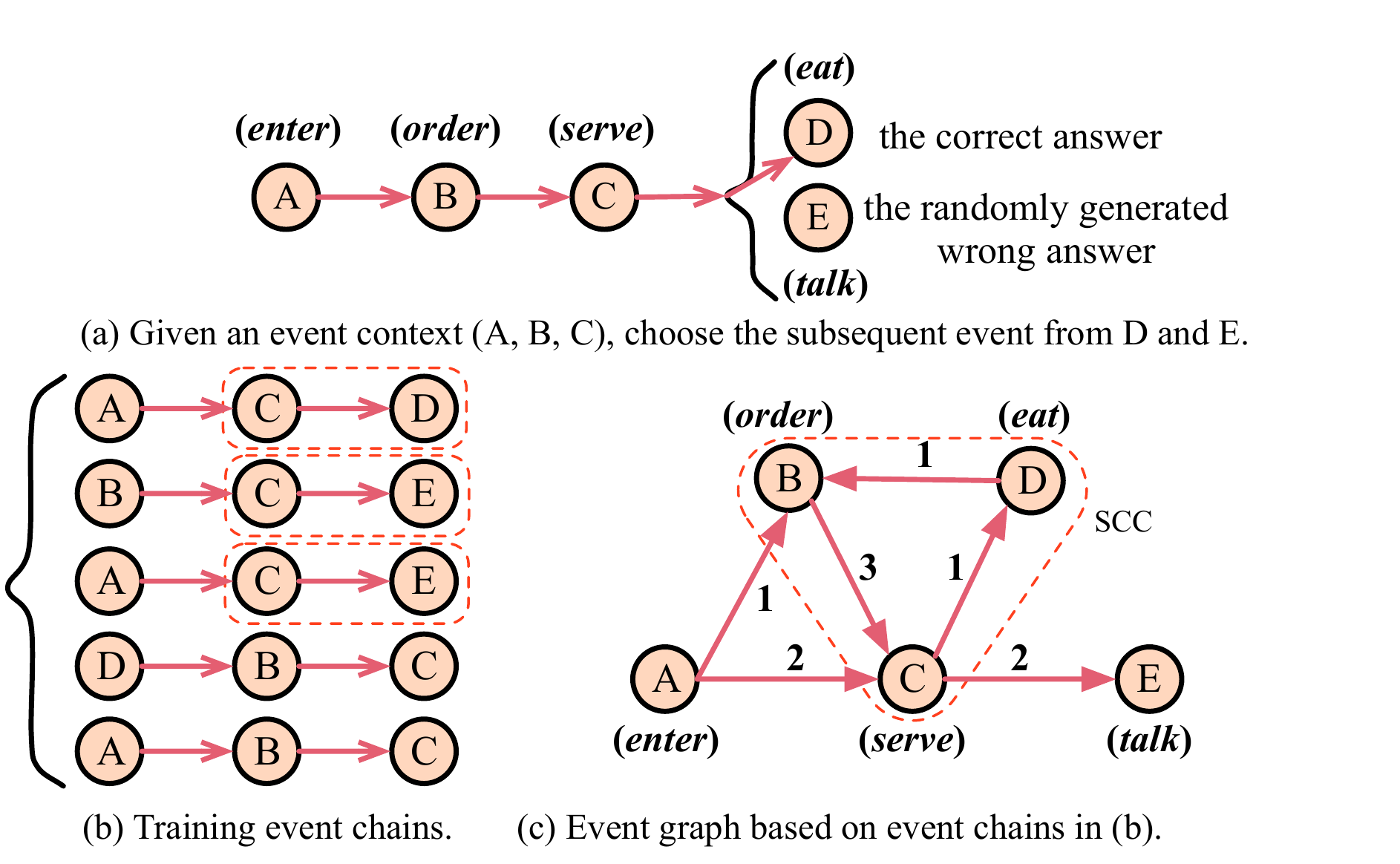}
    \vspace{-0.3cm}
    \caption{In (b), the training event chains show that C and E have a stronger relation than C and D, therefore event pair-based and chain-based models are very likely to choose the wrong, random candidate E. In (c), events B, C, and D compose a strongly connected component. This special graph structure contains dense connections information, which can help learn better event representations for choosing the correct subsequent event D. }
    \label{fig:example}
\end{figure}

Figure 2(a) gives an example to motive our idea of using more broader event connections (say graph structure). Given an event context A(\textit{enter}), B(\textit{order}), C(\textit{serve}), we need to choose the most reasonable subsequent event from the candidate list D(\textit{eat}) and E(\textit{talk}), where D(\textit{eat}) is the correct answer and E(\textit{talk}) is a randomly selected candidate event that occurs frequently in various scenarios. Pair-based and chain-based models trained on event chains datasets (as shown in Figure \ref{fig:example}(b)) are very likely to choose the wrong answer E, as training data show that C and E have a stronger relation than C and D. As shown in Figure \ref{fig:example}(c), by constructing an event graph based on training event chains, context events B, C and the candidate event D compose a strongly connected component, which indicates that D is a more reasonable subsequent event, given context events A, B, C.

Abstract event evolutionary principles and patterns are valuable commonsense knowledge, which is crucial for understanding narrative text, human behavior and social development. We use the notion of \textit{event evolutionary graph} (EEG) to denote the knowledge base that stores this kind of knowledge. Structurally, EEG is a directed cyclic graph, whose nodes are events and edges stand for the relations between events, e.g. temporal and causal relations. In this paper, we construct an event evolutionary graph based on narrative event chains, which is called narrative event evolutionary graph (NEEG). Having a NEEG in hand, another challenging problem is how to infer the subsequent event on the graph. A possible solution is to learn event representations based on network embedding. 

\citeauthor{duvenaud2015convolutional} \shortcite{duvenaud2015convolutional} introduced a convolutional neural network that could operate directly on graphs, which could be used for end-to-end learning of prediction tasks whose inputs are graphs of arbitrary size and shape. \citeauthor{kipf2016semi} \shortcite{kipf2016semi} presented a scalable approach for semi-supervised learning on graphs that was based on an efficient variant of convolutional neural networks. They chose a localized first-order approximation of spectral graph convolutions as the convolutional architecture, to scale linearly in the number of graph edges and learn hidden layer representations that encode both local graph structure and features of nodes. However, their models require the adjacency matrix to be symmetric and can only operate on undirected graphs. \citeauthor{gori2005new} \shortcite{gori2005new} proposed graph neural network (GNN), which extended recursive neural networks and could be applied on most of the practically useful kinds of graphs, including directed, undirected, labeled and cyclic graphs. However, the learning algorithm in their model required running the propagation to convergence, which could have trouble propagating information across a long range in a graph. To remedy this, \citeauthor{li2015gated} \shortcite{li2015gated} introduced modern optimization techniques of gated recurrent units to GNN. Nevertheless, their models can only operate on small graphs. 
In this paper, we further extend the work of \citeauthor{li2015gated} \shortcite{li2015gated} by proposing a scaled graph neural network (SGNN), which is feasible to large-scale graphs. We borrow the idea of divide and conquer in the training process that instead of computing the representations on the whole graph, SGNN processes only the concerned nodes each time.
By comparing between context event representations and candidate event representations learned from SGNN, we can choose the correct subsequent event.

This paper makes the following two key contributions:
\begin{itemize}[leftmargin=*]
\item We are among the first to propose constructing event graph instead of event pairs and event chains for the task of script event prediction.
\item We present a scaled graph neural network, which can model event interactions on large-scale dense directed graphs and learn better event representations for prediction.
\end{itemize}

Empirical results on widely used New York Times corpus show that our model achieves the best performance compared to state-of-the-art baseline methods, by using standard multiple choice narrative cloze (MCNC) evaluation. The data and code are released at \url{https://github.com/eecrazy/ConstructingNEEG_IJCAI_2018}.

\section{Model}

\begin{figure*} [th]
    \centering
    \includegraphics[width=1.8\columnwidth]{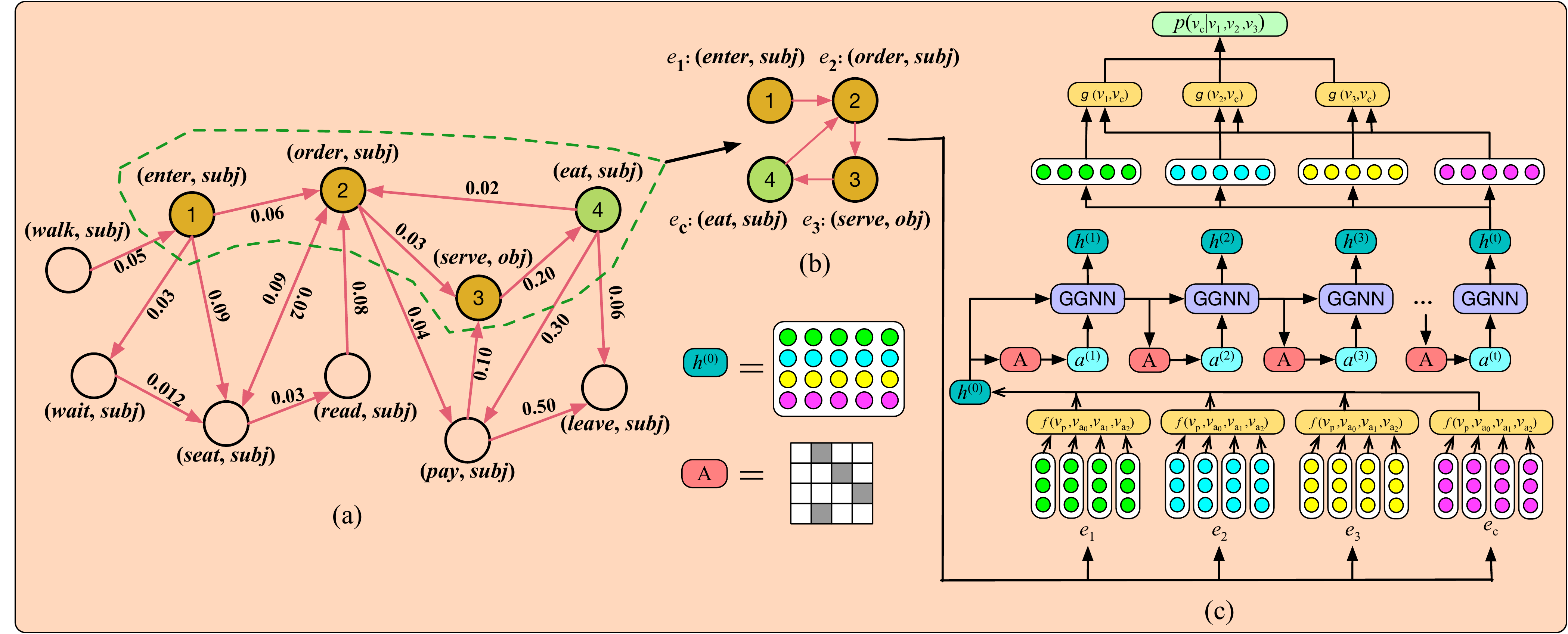}
    \caption{Framework of our proposed SGNN model. Suppose (a) is our constructed NEEG. Each time, a subgraph that contains all the context events (node 1,2,3) and all the corresponding candidate events (here we only draw one candidate event which is node 4) is chosen. The initial hidden representations $h^{(0)}$ and the adjacency matrix $\mathbf{A}$ are fed into GGNN to get the final event representations $h^{(t)}$, which are used to calculate the relatedness scores between the context and candidate events.}
    \label{fig:model}
\end{figure*}

As shown in Figure \ref{fig:model}, our model consists of two steps. The first step is to construct an event evolutionary graph based on narrative event chains. Second, we present a scaled graph neural network to solve the inference problem on the constructed event graph.

\subsection{Narrative Event Evolutionary Graph Construction}

NEEG construction consists of two steps: (1) we extract narrative event chains from newswire corpus; (2) construct NEEG based on the extracted event chains.

In order to compare with previous work, we adopt the same news corpus and event chains extraction methods as \cite{MarkGW-AAAI16}. We extract a set of narrative event chains $S=\{s_1,s_2,s_3,...,s_N\}$, where $s_i=\{T,e_1,e_2,e_3,...,e_m\}$. For example, $s_i$ can be \{$T$ = customer, walk($T$, restaurant, -), seat($T$, -, -), read($T$, menu, -), order($T$, food, -), serve(waiter, food, $T$), eat($T$, food, fork)\}. $T$ is the protagonist entity shared by all the events in this chain. $e_i$ is an event that consists of four components $\{p(a_0,a_1,a_2)\}$, where $p$ is the predicate verb, and $a_0,a_1,a_2$ are the subject, object and indirect object to the verb, respectively.

NEEG can be formally denoted as $G=\{V,E\}$, where $V=\{\mathbf{v}_1,\mathbf{v}_2,\mathbf{v}_3,...,\mathbf{v}_P\}$ is the node set, and $E=\{l_1,l_2,l_3,...,l_Q\}$ is the edge set. In order to overcome the sparsity problem of events, we represent event $e_i$ by its abstract form ($v_i$, $r_i$), where $v_i$ is denoted by a non-lemmatized predicate verb, and $r_i$ is the grammatical dependency relation of $v_i$ to the chain entity $T$, for example $e_i$=(\textit{eats, subj}). This kind of event representation is called predicate-GR \cite{MarkGW-AAAI16}. We count all the predicate-GR bigrams in the training event chains, and regard each predicate-GR bigram as an edge $l_i$ in $E$. Each $l_i$ is a directed edge $\mathbf{v}_i \rightarrow \mathbf{v}_j$ along with a weight $w$, which can be computed by:
\begin{equation} \small
w(\mathbf{v}_j|\mathbf{v}_i)=\frac {\text{count}(\mathbf{v}_i,\mathbf{v}_j)} {\sum_k \text{count}(\mathbf{v}_i,\mathbf{v}_k)}
\end{equation}
where $\text{count}(\mathbf{v}_i,\mathbf{v}_j)$ means the frequency of the bigram ($\mathbf{v}_i, \mathbf{v}_j$) appears in the training event chains.

The constructed NEEG $G$ has 104,940 predicate-GR nodes, and 6,187,046 directed and weighted edges. Figure \ref{fig:model}(a) illustrates a local subgraph in $G$, which describes the possible events involved in the restaurant scenario. Unlikely event pairs or event chains, event graph has dense connections among events and contains more abundant event interactions information.

\subsection{Scaled Graph Neural Network}
GNN was first proposed by \citeauthor{gori2005new} \shortcite{gori2005new}. \citeauthor{li2015gated} \shortcite{li2015gated} further introduced modern optimization technique of backpropagation through time and gated recurrent units to GNN, which is called gated graph neural network (GGNN). Nevertheless, GGNN needs to take the whole graph as inputs, thus it cannot effectively handle large-scale graph with hundreds of thousands of nodes. For the purpose of scaling to large-scale graphs, we borrow the idea of divide and conquer in the training process that we do not feed the whole graph into GGNN. Instead, only a subgraph (as shown in Figure \ref{fig:model}(b)) with context and candidate event nodes is fed into it for each training instance. Finally, the learned node representations can be used to solve the inference problem on the graph.

As shown in Figure \ref{fig:model}(c), the overall framework of SGNN has three main components. The first part is a representation layer, which is used to learn the initial event representation. The second part is a gated graph neural network, which is used to model the interactions among events and update the initial event representations. The third part is used to compute the relatedness scores between context and candidate events, according to which we can choose the correct subsequent event.

\subsubsection{Learning Initial Event Representations}

We learn the initial event representation by composing pretrained word embeddings of its verb and arguments. For arguments that consist of more than one word, we follow \cite{MarkGW-AAAI16} and only use the head word identified by the parser. Out-of-vocabulary words and absent arguments are represented by zero vectors.

Given an event $e_i=\{p(a_0,a_1,a_2)\}$ and the word embeddings of its verb and arguments $v_p,v_{a_0},v_{a_1},v_{a_2} \in \mathbb{R}^d$ ($d$ is the dimension of embeddings), there are several ways to get the representation of the whole event $v_{e_i}$ by a mapping function $v_{e_i}=f(v_p,v_{a_0},v_{a_1},v_{a_2})$. Here, we introduce three widely used semantic composition methods:

\begin{itemize}[leftmargin=*]
\item  \textbf{Average:} Use the mean value of the verb and all arguments vectors as the representation of the whole event.
\item  \textbf{Nonlinear Transformation} \cite{wang2017integrating}:
\begin{equation} \small
v_e=\tanh(W_p\bm\cdot v_p+W_0\bm\cdot v_{a_0}+W_1\bm\cdot v_{a_1}+W_2\bm\cdot v_{a_2}+b)
\end{equation}
where $W_p,W_0,W_1,W_2,b$ are model parameters.
\item  \textbf{Concatenation} \cite{MarkGW-AAAI16}: Concatenate the verb and all argument vectors as the representation of the whole event.
\end{itemize}

\subsubsection{Updating Event Representations Based on GGNN}

As introduced above, GGNN is used to model the interactions among all context and candidate events. The main challenge is how to train it on a large-scale graph. To train the GGNN model on NEEG with more than one hundred thousand event nodes, each time we feed into it a small subgraph, instead of the whole graph, to make it feasible to large-scale graphs. 

Inputs to GGNN are two matrices $h^{(0)}$ and $\mathbf{A}$, where $h^{(0)}$=$\{v_{e_1},v_{e_2},...,v_{e_n},v_{e_{c_1}},v_{e_{c_2}},...,v_{e_{c_k}}\}$ ($n$ is 8 and $k$ is 5, the same as \cite{MarkGW-AAAI16}), contains the initial context and subsequent candidate event vectors, and $\mathbf{A}$$\in \mathbb{R}^{(n+k) \times (n+k)}$ is the corresponding subgraph adjacency matrix, here:

\begin{equation} \small
\mathbf{A}\left[i,j\right] =
\begin{cases} 
w(\mathbf{v}_j|\mathbf{v}_i),  & \mbox{if }\mathbf{v}_i \rightarrow \mathbf{v}_j \in E, \\
0, & \mbox{others}.
\end{cases}    
\end{equation}
The adjacency matrix $\mathbf{A}$ determines how nodes in the subgraph interact with each other.
The basic recurrence of GGNN is:

\begin{align}
a^{(t)} & = \mathbf{A}^\top h^{(t-1)}+ b \\
z^{(t)}&= \sigma(W^za^{(t)}+U^zh^{(t-1)}) \\
r^{(t)} &= \sigma(W^ra^{(t)}+U^rh^{(t-1)}) \\
c^{(t)} &= \tanh(Wa^{(t)}+U(r^{(t)}\odot h^{(t-1)})) \\
h^{(t)} &= (1-z^{(t)})\odot h^{(t-1)}+z^{(t)}\odot c^{(t)}
\end{align}

GGNN behaves like the widely used gated recurrent unit (GRU) \cite{cho2014learning}. Eq. (4) is the step that passes information between different nodes of the graph via directed adjacency matrix $\mathbf{A}$. $a^{(t)}$ contains activations from edges. The remainings are GRU-like updates that incorporate information from the other nodes and from the previous time step to update each node's hidden state. $z^{(t)}$ and $r^{(t)}$ are the update and reset gate, $\sigma$ is the logistic sigmoid function, and $\odot$ is element-wise multiplication.
We unroll the above recurrent propagation for a fixed number of steps $\mathbf{K}$. The output ${h}^{(t)}$ of GGNN can be used as the updated representations of context and candidate events.

\subsubsection{Choosing the Correct Subsequent Event}

After obtaining the hidden states for each event, we model event pair relations using these hidden state vectors. A straightforward approach to model the relation between two events is using a Siamese network \cite{MarkGW-AAAI16}. The output of GGNN for context events are $h^{(t)}_1,h^{(t)}_2,...,h^{(t)}_n$ and for the candidate events are $h^{(t)}_{c_1},h^{(t)}_{c_2},...,h^{(t)}_{c_k}$. Given a pair of events $h^{(t)}_i$ ($i \in \left[1...n\right]$) and $h^{(t)}_{c_j}$ ($j \in \left[1...k\right]$), the relatedness score is calculated by $s_{ij}=g(h^{(t)}_i,h^{(t)}_{c_j})$, where $g$ is the score function.

There are multiple choices for the score function $g$ in our model. Here, we introduce four common used similarity computing metrics that can serve as $g$ in the followings.

\begin{itemize}[leftmargin=*]
\item \textbf{Manhattan Similarity} is the Manhattan distance of two vectors: $manhattan(h^{(t)}_i,h^{(t)}_{c_j})= \sum |h^{(t)}_i-h^{(t)}_{c_j}|$.
\item \textbf{Cosine Similarity} is the cosine distance of two vectors: $cosine(h^{(t)}_i,h^{(t)}_{c_j})=\frac {h^{(t)}_i \bm\cdot h^{(t)}_{c_j}} {||h^{(t)}_i||||h^{(t)}_{c_j}||}$.
\item \textbf{Dot Similarity} is the inner product of two vectors: $dot(h^{(t)}_i,h^{(t)}_{c_j})= h^{(t)}_i \bm\cdot h^{(t)}_{c_j}$.
\item \textbf{Euclidean Similarity} is the euclidean distance of two vectors: $euclid(h^{(t)}_i,h^{(t)}_{c_j})= ||h^{(t)}_i-h^{(t)}_{c_j}||$.
\end{itemize}

Given the relatedness score $s_{ij}$ between each context event $h^{(t)}_i$ and each subsequent candidate event $h^{(t)}_{c_j}$, the likelihood of $e_{c_j}$ given $e_1,e_2,...,e_n$ can be calculated as $s_j=\frac 1 n \sum_{i=1}^ns_{ij}$, then we choose the correct subsequent event by $c=\max_{j}{s_j}$.

We also use the attention mechanism \cite{Bahdanau2014Neural} to the context events, as we believe that different context events may have different weight for choosing the correct subsequent event. We use an attentional neural network to calculate the relative importance of each context event according to the subsequent event candidates:

\begin{align}
u_{ij}&=\tanh(W_hh^{(t)}_i+W_ch^{(t)}_{c_j}+b_u)\\
\alpha_{ij}&=\frac {\exp(u_{ij})} {\sum_k \exp(u_{kj})}
\end{align}
Then the relatedness score is calculated by:
\begin{equation} \small
s_{ij}=\alpha_{ij}g(h^{(t)}_i,h^{(t)}_{c_j})
\end{equation}

\subsubsection{Training Details}

All the hyperparameters are tuned on the development set, and we use margin loss as the objective function:
\begin{equation*} \small
L(\Theta) = \sum_{I=1}^N \sum_{j=1}^k(\max(0, margin - s_{Iy} + s_{Ij})) +\frac \lambda 2 ||\Theta||^2
\end{equation*}
where $s_{Ij}$ is the relatedness score between the $I$th event context and the corresponding $j$th subsequent candidate event, $y$ is the index of the correct subsequent event. The $margin$ is the margin loss function parameter, which is set to 0.015. $\Theta$ is the set of model parameters. $\lambda$ is the parameter for L2 regularization, which is set to 0.00001. The learning rate is 0.0001, batch size is 1000, and recurrent times $\mathbf{K}$ is 2. 

We use DeepWalk algorithm \cite{perozzi2014deepwalk} to train embeddings for predicate-GR on the constructed NEEG (we find that embeddings trained from DeepWalk on the graph are better than that from Word2vec trained on event chains), and use the Skipgram algorithm \cite{mikolov2013distributed} to train embeddings for arguments $a_0,a_1,a_2$ on event chains. The embedding dimension $d$ is 128. The model parameters are optimized by the RMSprop algorithm. Early stopping is used to judge when to stop the training loop.

\section{Evaluation}

We evaluate the effectiveness of SGNN comparing with several state-of-the-art baseline methods. Accuracy (\%) of choosing the correct subsequent event is used as the evaluation metric.

\subsection{Baselines}
We compare our model with the following baseline methods.
\begin{itemize}[leftmargin=*]
\item \textbf{PMI} \cite{chambers2008} is the co-occurrence-based model that calculates predicate-GR event pairs relations based on Pairwise Mutual Information.
\item \textbf{Bigram} \cite{jans2012skip} is the counting-based skip-grams model that calculates event pair relations based on bigram probabilities.
\item \textbf{Word2vec} \cite{mikolov2013distributed} is the widely used model that learns word embeddings from large-scale text corpora. The learned embeddings for verbs and arguments are used to compute pairwise event relatedness scores.
\item \textbf{DeepWalk} \cite{perozzi2014deepwalk} is the unsupervised model that extends the word2vec algorithm to learn embeddings for networks.
\item \textbf{EventComp} \cite{MarkGW-AAAI16} is the neural network model that simultaneously learns embeddings for the event verb and arguments, a function to compose the embeddings into a representation of the event, and a coherence function to predict the strength of association between two events.
\item \textbf{PairLSTM} \cite{wang2017integrating} is the model that integrates event order information and pairwise event relations together by calculating pairwise event relatedness scores using the LSTM hidden states as event representations.
\end{itemize}

\subsection{Dataset}

\begin{table} \small
    \centering
    \begin{tabular} {l c c c}
        \toprule
        &\textbf{Training}& \textbf{Development}& \textbf{Test}  \\
        \midrule
        \#Documents & 830,643 & 103,583 & 103,805  \\
        \#Chains for NEEG  & 5,997,385 & - & - \\
        \#Chains for SGNN & 140,331 & 10,000 & 10,000 \\
       \bottomrule
    \end{tabular}
    \vspace{-0.2cm}
    \caption{Statistics of our datasets.}
    \label{tab:data}
\end{table}

Following \citeauthor{MarkGW-AAAI16} \shortcite{MarkGW-AAAI16}, we extract event chains from the New York Times portion of the Gigaword corpus. The C\&C tools \cite{curran2007linguistically} are used for POS tagging and dependency parsing, and OpenNLP is used for phrase structure parsing and coreference resolution. There are 5 candidate subsequent events for each event context and only one of them is correct. The detailed dataset statistics are shown in Table \ref{tab:data}.

\section{Results and Analysis}

\subsection{Overall Results}
\begin{table}[t] \small
    \centering
    \begin{tabular} {l p{1.2cm}<{\centering}} 
        \toprule
        \textbf{Methods}& \textbf{Accuracy}  \\
        \midrule
        Random & 20.00 \\
        PMI \cite{chambers2008} & 30.52 \\
        Bigram \cite{jans2012skip}& 29.67 \\
        Word2vec \cite{mikolov2013distributed}& 42.23 \\
        DeepWalk  \cite{perozzi2014deepwalk} & 43.01 \\
        EventComp \cite{MarkGW-AAAI16}& 49.57  \\
        PairLSTM \cite{wang2017integrating}& 50.83 \\
        \midrule
        SGNN-attention (without attention) & 51.56  \\
        SGNN (ours) & \textbf{52.45}  \\
        SGNN+PairLSTM & 52.71  \\
        SGNN+EventComp & 54.15  \\
        SGNN+EventComp+PairLSTM & \textbf{54.93} \\
        \bottomrule
    \end{tabular}
    \caption{Results of script event prediction accuracy (\%) on the test set. SGNN-attention is the SGNN model without attention mechanism. Differences between SGNN and all baseline methods are significant ($p < 0.01$) using t-test, except SGNN and PairLSTM ($p = 0.246$).}
    \label{tab:final}
\end{table}

\begin{figure} [t] 
    \centering
    \includegraphics[width=0.8\columnwidth]{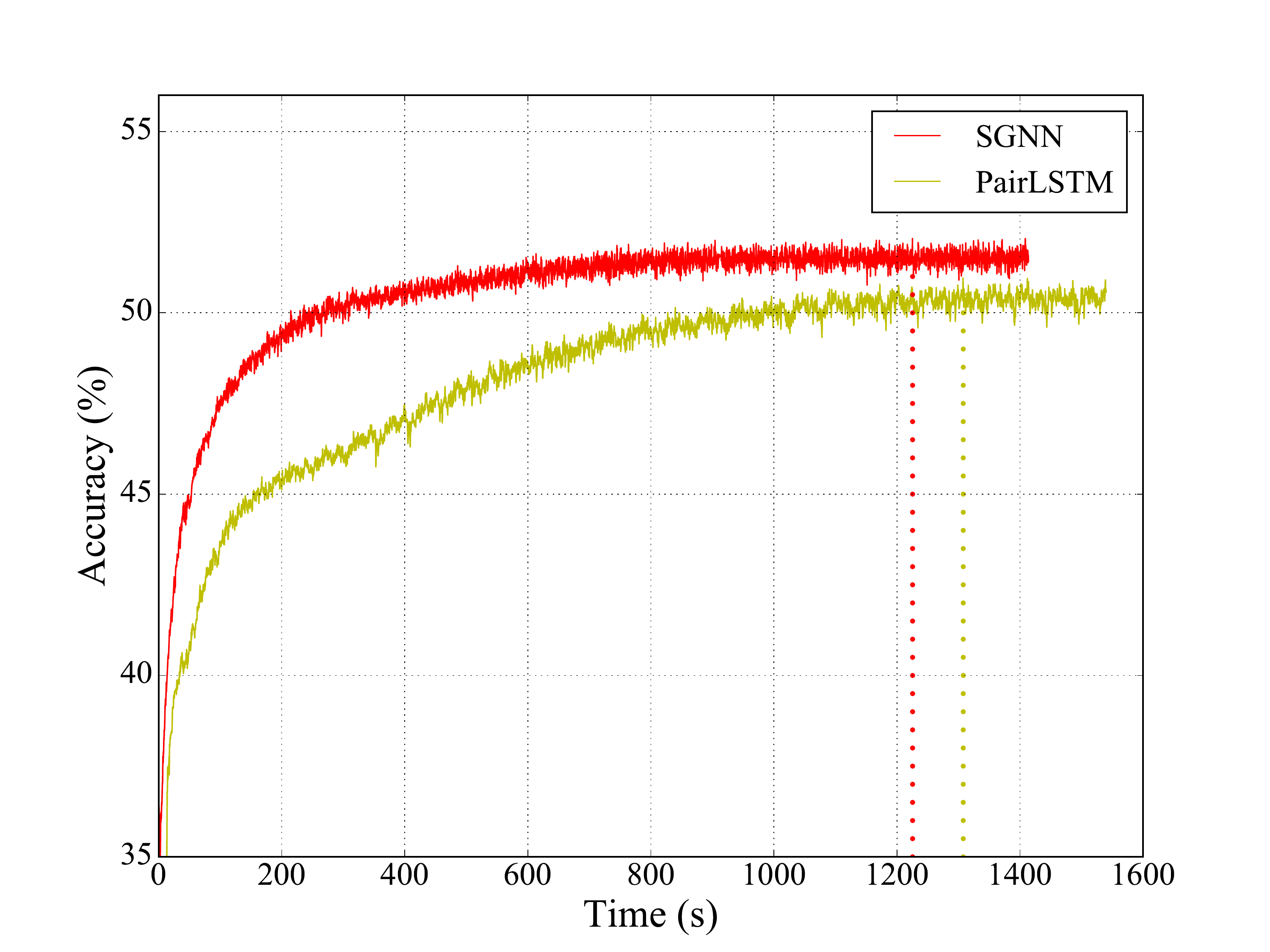}
    \vspace{-0.3cm}
    \caption{Learning curves on development set of SGNN and PairLSTM models, using the same learning rate and batch size.}
    \label{fig:time}
\end{figure}

Experimental results are shown in Table \ref{tab:final}, from which we can make the following observations.

(1) Word2vec, DeepWalk and other neural network-based models (EventComp, PairLSTM, SGNN) achieve significantly better results than the counting-based PMI and Bigram models. The main reason is that learning low dimensional dense embeddings for events is more effective than sparse feature representations for script event prediction.

(2) Comparison between ``Word2vec'' and ``DeepWalk'', and between ``EventComp, PairLSTM'' and ``SGNN'' show that graph-based models achieve better performance than pair-based and chain-based models. This confirms our assumption that the event graph structure is more effective than event pairs and chains, and can provide more abundant event interactions information for script event prediction.

(3) Comparison between ``SGNN-attention'' and ``SGNN'' shows the attention mechanism can effectively improve the performance of SGNN. This indicates that different context events have different significance for choosing the correct subsequent event.

(4) SGNN achieves the best script event prediction performance of 52.45\%, which is 3.2\% improvement over the best baseline model (PairLSTM).

We also experimented with combinations of different models, to observe whether different models have complementary effects to each other. We find that SGNN+EventComp boosts the SGNN performance from 52.45\% to 54.15\%. This shows that they can benefit from each other. Nevertheless, SGNN+PairLSTM can only boost the SGNN performance from 52.45\% to 52.71\%. This is because the difference between SGNN and PairLSTM is not significant, which shows that they may learn similar event representations but SGNN learns in a better way. The combination of SGNN, EventComp and PairLSTM achieves the best performance of 54.93\%. This is mainly because pair structure, chain structure and graph structure each has its own advantages and they can complement each other.

The learning curve (accuracy with time) of SGNN and PairLSTM is shown in Figure \ref{fig:time}. We find that SGNN quickly reaches a stable high accuracy, and outperforms PairLSTM from start to the end. This demonstrates the advantages of SGNN over PairLSTM model.

\subsection{Comparative Experiments}
We conduct several comparative experiments on the development set to study the influence of various settings on SGNN.

\subsubsection{Experiment with Different Event Semantic Composition Methods}

\begin{table} [t] \small
    \centering
    \begin{tabular} {l c}
        \toprule
        \textbf{Composition}& \textbf{Accuracy (\%)}  \\
        \midrule
        Average & 43.42  \\
        Nonlinear & 51.54 \\
        Concatenation & \textbf{52.38} \\
       \bottomrule
    \end{tabular}
    \caption{Influence of event arguments composition approaches.}
    \label{tab:composition}
\end{table}

\begin{table} [t]\small
    \centering
    \begin{tabular} {l c} 
        \toprule
        \textbf{Score Metric}& \textbf{Accuracy (\%)}  \\
        \midrule
        Manhattan & 50.11 \\
        Cosine & 50.81  \\
        Dot & 51.62  \\
        Euclidean & \textbf{52.38}  \\
       \bottomrule
    \end{tabular}
    \caption{Influence of similarity score metrics.}
    \label{tab:metric}
\end{table}
Given the word embeddings of the verb and arguments $v_p$,$v_{a_0}$,$v_{a_1}$,$v_{a_2}$ of the event $e_i$, we compare three event semantic composition methods, introduced in Section 2.2.1. Experimental results are shown in Table \ref{tab:composition}.

We find that concatenating the embeddings of the verb and arguments vectors can achieve the best performance. And average input vectors is the worst way to get the representation $v_e$. The main reason is that many events do not have an indirect object $a_2$, which may harm the performance of the averaging operation. 

\subsubsection{Experiment with Different Score Functions}
We compare several common used similarity score metrics $g$, as introduced in Section 2.2.3, to investigate their influence on the performance. As shown in Table \ref{tab:metric}, we find that different score metrics indeed have different effects on the performance, though the gaps among them are not so big. Euclidean score metric achieves the best result, which is consistent with the results of the previous study using word embeddings to compute document distances \cite{kusner2015word}.


\section{Related Work}

\subsection{Statistical Script Learning}
The use of scripts in AI dates back to the 1970s~\cite{schank1977scripts}. In this conception, \emph{scripts} are composed of complex events without probabilistic semantics. In recent years, a growing body of research has investigated learning probabilistic co-occurrence-based models with simpler events. \citeauthor{chambers2008} \shortcite{chambers2008} proposed unsupervised induction of \textit{narrative event chains} from raw newswire text, with \textit{narrative cloze} as the evaluation metric, and pioneered the recent line of work on statistical script learning. \citeauthor{jans2012skip} \shortcite{jans2012skip} used bigram model to explicitly model the temporal order of event pairs. 
However, they all utilized a very impoverished representation of events in the form of (\textit{verb, dependency}). To overcome the drawback of this event representation, \citeauthor{pichotta2014statistical} \shortcite{pichotta2014statistical} presented an approach that employed events with multiple arguments. 

There have been a number of recent neural models for script learning. \citeauthor{pichotta2015learning} \shortcite{pichotta2015learning} showed that LSTM-based event sequence model outperformed previous co-occurrence-based methods. \citeauthor{pichotta2016using} \shortcite{pichotta2016using} used a Seq2Seq model directly operating on raw tokens to predict sentences. \citeauthor{MarkGW-AAAI16} \shortcite{MarkGW-AAAI16} described a feedforward neural network which composed verbs and arguments into low-dimensional vectors, evaluating on a multiple-choice version of the narrative cloze task. \citeauthor{wang2017integrating} \shortcite{wang2017integrating} integrated event order information and pairwise event relations together by calculating pairwise event relatedness scores using the LSTM hidden states. 

Previous studies built prediction models either based on event pairs \cite{chambers2008,MarkGW-AAAI16} or based on event chains \cite{wang2017integrating}. In this paper, we propose to solve the problem of script event prediction based on event graph structure.

\subsection{Neural Network on Graphs}
Graph-structured data appears frequently in domains such as social networks and knowledge bases.
\cite{perozzi2014deepwalk} proposed the unsupervised DeepWalk algorithm that extended the word2vec \cite{mikolov2013distributed} algorithm to learn embeddings for graph nodes based on random walks. Later, unsupervised network embedding algorithms including LINE \cite{Tang2015LINE} and node2vec \cite{Grover2016node2vec} have been proposed following DeepWalk.~\citeauthor{duvenaud2015convolutional} \shortcite{duvenaud2015convolutional} introduced a convolutional neural network that could operate directly on graphs.~\citeauthor{kipf2016semi} \shortcite{kipf2016semi} presented a scalable approach for semi-supervised learning on graphs that was based on an efficient variant of convolutional neural networks. However, their models require the adjacency matrix to be symmetric and can only operate on undirected graphs.


Most related to our SGNN model is the graph neural network introduced by \citeauthor{gori2005new} \shortcite{gori2005new}, which was capable of directly processing graphs. GNN extended recursive neural networks and could be applied on most of the practically useful kinds of graphs, including directed, undirected, labeled and cyclic graphs. However, the learning algorithm they used required running the propagation to convergence, which could have trouble propagating information across a long range in a graph. To remedy this, \citeauthor{li2015gated} \shortcite{li2015gated} introduced modern optimization technique of backpropagation through time and gated recurrent units to GNN, resulting in the gated graph neural network model. Nevertheless, their models usually operate on small graphs. 

In this paper, we further extend the GGNN model by proposing a scaled graph neural network, which is feasible to large-scale graphs.
Instead of feeding the whole graph into the model, we borrow the idea of divide and conquer in the training process, and only a subgraph that contains the concerned nodes is fed into GGNN for each training instance.
\subsection{Graph-based Organization of Events}

Some previous studies also explored graph-based organization of events.~\citeauthor{Orkin2007Learning} \shortcite{Orkin2007Learning} described events in a multiagent script, which were derived from data collected from video game players.~\citeauthor{Li2013Story} \shortcite{Li2013Story} described an approach to construct graph-based narrative scripts using crowdsourcing for story generation.~\citeauthor{Glava2015Construction} \shortcite{Glava2015Construction} proposed \textit{event graphs} as a novel way of structuring event-based information from text. Other studies tried to organize temporal relations~\cite{chambers2008} or causal relations~\cite{zhao2017constructing} between events into graph structure. In this paper, we propose \textit{event evolutionary graph}, which denotes the knowledge base storing abstract event evolutionary patterns.

\section{Conclusion}

In this paper, we propose constructing event graph to solve the script event prediction problem based on network embedding. In particular, to better utilize the dense connections information among events, we construct a narrative event evolutionary graph (NEEG) based on the extracted narrative event chains. To solve the inference problem on NEEG, we present a scaled graph neural network (SGNN) to model the events interactions and learn better event representations for choosing the correct subsequent event. Experimental results show that event graph structure is more effective than event pairs and event chains, which can help significantly boost the prediction performance and make the model more robust.

\section*{Acknowledgments}

This work is supported by the National Key Basic Research Program of China via grant 2014CB340503 and the National Natural Science Foundation of China (NSFC) via grants 61472107 and 61702137. The authors would like to thank the anonymous reviewers for the insightful comments. They also thank Haochen Chen and Yijia Liu for the helpful discussion.

\bibliographystyle{named}
\bibliography{ijcai18}
\end{document}